\documentclass[10pt,twocolumn,letterpaper]{article}

\usepackage{wrap}
\usepackage{booktabs}
\usepackage{colortbl}
\usepackage{arydshln}
\usepackage{multirow}
\usepackage{graphicx}
\usepackage{url}
\usepackage{enumitem}
\usepackage{pifont}
\usepackage{wrapfig}
\usepackage{xcolor}
\usepackage{fontawesome5}
\usepackage{tikz}

\newcommand{\itbf}[1]{\textit{\textbf{#1}}}
\newcommand{\tablerowcolor}{\rowcolor{gray!15}}
\definecolor{wrapblue}{rgb}{0.21,0.49,0.74}
\definecolor{deepblue}{RGB}{1,0,122}
\usepackage[pagebackref,breaklinks,colorlinks,allcolors=wrapblue]{hyperref}

\def\confName{}
\def\confYear{2027}

\title{WoRld As a Program: Programmable Spatial Reasoning for Multimodal LLMs}

\author{Wenbin Wang$^{1,4}$\hspace{0.7cm}Xianlong Liu$^{1}$\hspace{0.7cm}Liang Ding$^{2}$\hspace{0.7cm}Mengya Han$^{1}$\\Han Hu$^{3}$\hspace{0.7cm}Yong Luo$^{1\dagger}$\hspace{0.7cm}Bo Du$^{1}$\hspace{0.7cm}Pan Zhou$^{4\dagger}$\hspace{0.7cm}\\
$^{1}$Wuhan University~~~
$^{2}$Alibaba Group~~~\\
$^{3}$Beijing Institute of Technology~~~
$^{4}$Singapore Management University\\[8pt]
\makebox[\textwidth][c]{%
    wangwenbin97@whu.edu.cn%
}
}

\begin{document}
\maketitle

\begingroup
\renewcommand{\thefootnote}{}
\footnotetext{$\dagger$ Corresponding authors.}
\endgroup

\begin{abstract}
Although Multimodal Large Language Models (MLLMs) have advanced rapidly, spatial reasoning remains limited by implicit, geometrically unconstrained representations. Meanwhile, existing explicit-geometry approaches may produce inaccurate or incomplete object geometry, limiting the reliability of subsequent spatial reasoning. 
Recent advances in 3D reconstruction enable explicit scene representations from ordinary videos. This raises a natural question: \itbf{Can explicit 3D world states be made directly programmable for MLLMs?}
To this end, we propose \itbf{WoRld As a Program (WRAP)}, a framework that converts standard video observations into an addressable and executable 3D scene state for MLLMs, where visual objects are exposed as \textit{Grounded Variables} for \textit{3D Program Reasoning}. 
\itbf{WRAP} constructs candidate object point clouds from video observations and exposes them as Grounded Variables through persistent visual identifiers. The MLLM selects query-relevant variables from tagged frames and generates executable programs over their associated geometry.
Extensive experiments demonstrate the effectiveness of \itbf{WRAP}, which achieves SOTA performance on ReVSI with up to 5.2\% accuracy and outperforms prior methods on EditRoom-DB by up to 35.3\% IoU. These results highlight the benefits of programmable 3D representations for spatial reasoning and scene editing.

\end{abstract}

\noindent
\textcolor[RGB]{65,112,176}{\faGlobe}\ \textbf{Project Page:}
\href{https://dreammr.github.io/WoRld-As-a-Program/}{\textit{\textbf{W}o\textbf{R}ld-\textbf{A}s-a-\textbf{P}rogram}}

\section{Introduction}

\begin{figure}[t]
    \centering
    \includegraphics[width=1.\linewidth]{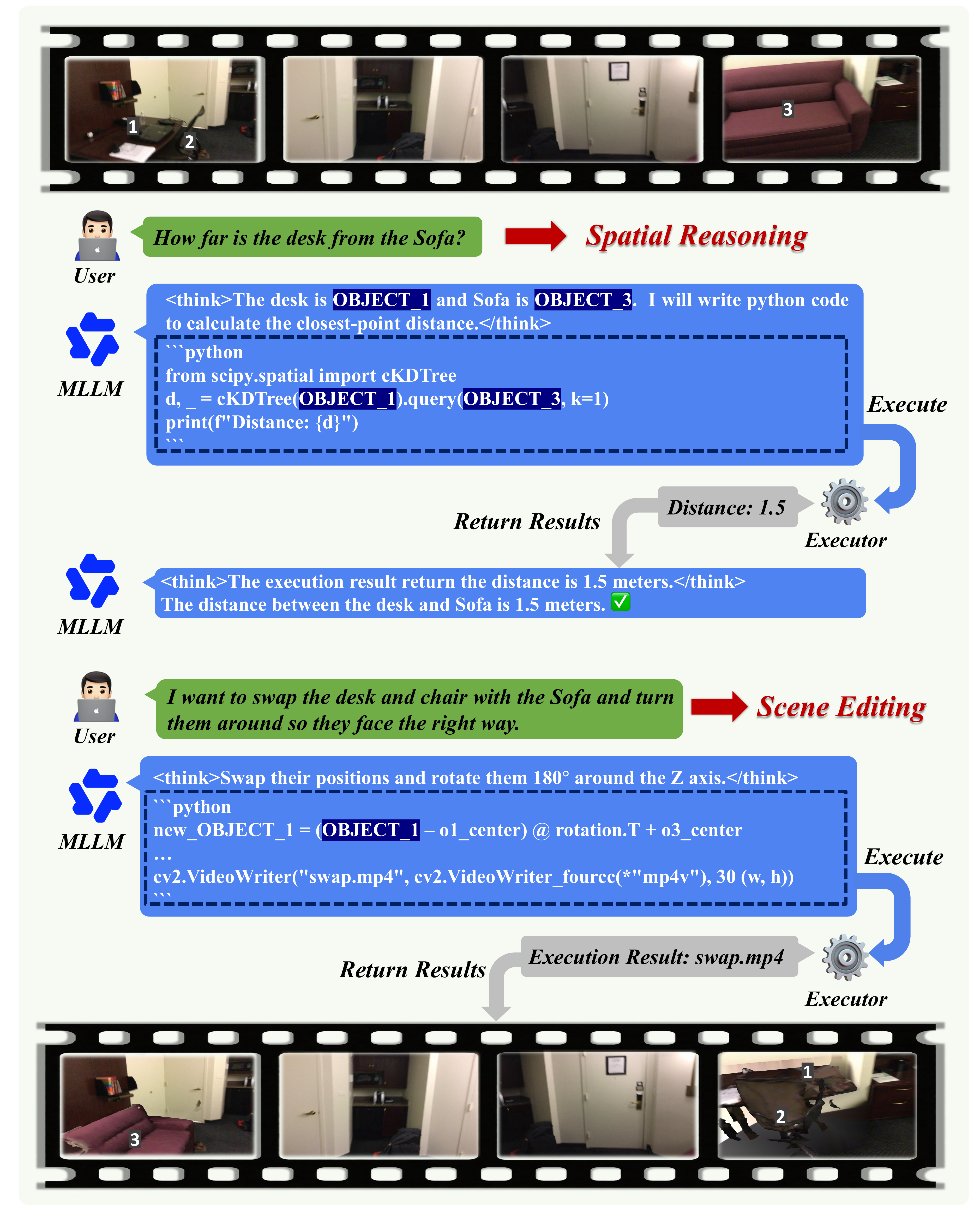}
    \caption{\itbf{Running example of \itbf{WRAP} on spatial reasoning and scene editing tasks.} \itbf{WRAP} directly identifies target objects from tagged videos and binds them to \colorbox{deepblue}{\textcolor{white}{\textbf{Grounded Variables}}}, each representing a 3D object point cloud. It then autonomously generates and executes task-specific code to complete the tasks.}
    \label{fig:motivation}
\end{figure}

Multimodal Large Language Models (MLLMs) have recently demonstrated impressive capabilities in perception and reasoning from images and videos~\cite{liu2024improved, Qwen-VL, feng2026video}. However, their spatial reasoning is still largely performed over implicit visual representations and autoregressive token generation, without explicitly enforcing the geometric structure underlying the observed scene. As a result, spatial evidence acquired from different views is not necessarily grounded in a single, globally consistent 3D representation, which can lead to failures in tasks that require metric estimation, viewpoint-invariant relations, or cross-view consistency. In this sense, current MLLMs may combine fragmented visual cues to produce plausible answers without necessarily constructing a coherent \textit{\texttt{spatial mental model}}~\cite{tversky1993cognitive} of the environment.

To address this limitation, recent studies have explored two major paradigms for enhancing spatial reasoning in MLLMs: \textit{Implicit Spatial Modeling} and \textit{Explicit Geometry Modeling}. Implicit spatial modeling methods~\cite{chen2026reasoning,fan2026vlm,zheng2026learning,chen2026think,yang2026visual,cui2026dual,gao2026map2thought} incorporate spatial knowledge into MLLMs through spatially enriched training data or dedicated spatial encoders, encouraging the models to learn implicit representations of spatial structure. However, such representations are not necessarily stable or globally consistent. For example, even on absolute-distance tasks that are independent of temporal order, simply reversing the temporal order of the input video frames can lead to substantial changes in model predictions. In contrast, explicit geometry modeling methods use external spatial tools to reconstruct 3D representations and support spatial reasoning. However, some methods~\cite{dai2026s, cho2026spatialclaw, marsili2025visual} intertwine query-conditioned object grounding by the MLLM with geometric acquisition, potentially constraining the independent refinement and reuse of object-level geometry. Moreover, some methods~\cite{dai2026s, chen2026spacetools} rely on a fixed set of tools, limiting their flexibility across diverse spatial tasks.

Recent advances in 3D reconstruction methods~\cite{worldlabs_marble, leroy2024grounding,wang2026vggt,wang2025vggt,wang2024dust3r} enable globally consistent scene geometry to be recovered from multi-view observations, making scene entities and their spatial relations accessible for explicit computation. This naturally raises a key question: \itbf{Can explicit 3D world states be made directly programmable by MLLMs?} To explore this question, we introduce \itbf{\underline{W}o\underline{R}ld \underline{A}s a \underline{P}rogram (WRAP)}, which connects video observations to an external 3D scene state and exposes grounded scene entities as programmable variables. Our study is organized around three questions:

\noindent\itbf{Q1: How can MLLMs bind objects in videos to addressable 3D entities?}
We address this through \textbf{Visual-Geometric Entity Binding}, which establishes explicit correspondences between visual object instances and their associated 3D point clouds. Object-level point clouds are first reconstructed from the input video and assigned persistent symbolic identifiers. During reasoning, the MLLM identifies query-relevant objects from the tagged video and selects their corresponding identifiers. This design decouples MLLM-based object grounding from 3D reconstruction, enabling direct identification of query-relevant objects from visual input.

\noindent\itbf{Q2: How can MLLMs formulate executable spatial programs over these grounded variables?}
We introduce \textbf{MLLM-Native Spatial Reasoning}, which enables the model to generate executable programs over grounded 3D entities. Given a query and tagged video, the MLLM identifies relevant object variables and flexibly constructs task-specific reasoning procedures as code. This enables computations beyond a fixed set of tools, while deterministic execution is delegated to an external executor.

\noindent\itbf{Q3: How effective is WRAP for spatial reasoning and scene editing?}
To answer Q3, we evaluate \itbf{WRAP} on video spatial reasoning and 3D scene editing tasks. \itbf{WRAP} achieves 66.1\% accuracy on ReVSI~\cite{zhang2026revsi}, surpassing the strongest reported baseline by 5.2 percentage points, and outperforms prior methods on EditRoom-DB~\cite{zheng2025editroom}. Further analysis highlights the importance of accurate object grounding for program-based spatial reasoning.

Our contributions can be summarized as:
\begin{itemize}[leftmargin=*, labelsep=0.5em]
    \item We propose \itbf{WRAP}, a spatial reasoning framework that connects video observations to an executable 3D scene state. By exposing scene entities as \textit{Grounded Variables}, \itbf{WRAP} provides access to explicit geometry while preserving the native video interface of the MLLM.
    \item We introduce \itbf{Visual-Geometric Entity Binding} to link visual objects to addressable 3D variables (\textit{Grounded Variables}), and \itbf{MLLM-Native Spatial Reasoning}  to generate executable programs over these variables, with deterministic computation delegated to an external executor.
    \item We construct a \textbf{$27$K training dataset} for 3D program reasoning, covering grounded variable understanding, adaptive 3D activation, and executable spatial program generation, and introduce a two-stage learning strategy to progressively acquire these capabilities. 
    \item We empirically demonstrate the effectiveness of \itbf{WRAP} across video spatial reasoning and 3D scene editing, achieving state-of-the-art performance with a relative improvement of \textbf{8.5\%} on ReVSI and an IoU improvement of up to \textbf{35.3\%} on EditRoom-DB test sets.
\end{itemize}

\section{Related Work} 
Recent approaches to spatial reasoning incorporate geometric priors and 3D structure through two paradigms: \textit{Implicit Spatial Modeling} and \textit{Explicit Geometry Modeling}.

\paragraph{Implicit Spatial Modeling Methods.}
These methods enhance spatial reasoning through geometric encoders or large-scale spatial supervision. Some methods~\cite{fan2026vlm,zheng2026learning,li2026thinking} integrate dedicated geometric representations or encoders into MLLMs, while others, such as Cambrian-S~\cite{yang2026cambrian} and SpaceR~\cite{ouyang2025spacer}, rely on spatially enriched training data and reasoning supervision. However, the resulting implicit representations may lack stability and global consistency. 

\paragraph{Explicit Geometry Modeling Methods.}
These approaches~\cite{marsili2025visual, dai2026s, luo2026pyspatial, chen2026thinking, yuan2026boosting} externalize spatial information into explicit 3D representations or executable geometric operations. VADAR~\cite{marsili2025visual} dynamically synthesizes spatial APIs, while S-Agent~\cite{dai2026s} and SpatialClaw~\cite{cho2026spatialclaw} perform tool- or code-based geometric reasoning. Accurate object grounding is essential for spatial reasoning, as subsequent geometric computations depend on identifying the correct objects. However, some existing approaches~\cite{dai2026s, cho2026spatialclaw, marsili2025visual} intertwine query-conditioned object grounding by the MLLM with geometric acquisition, potentially constraining the independent refinement and reuse of object-level geometry. Others~\cite{dai2026s, chen2026spacetools} rely on a fixed set of tools, which limits their flexibility in handling diverse spatial reasoning tasks.

Overall, prior methods may lack a globally consistent 3D state, intertwine object grounding with geometric acquisition, or rely on fixed tools. \itbf{WRAP} instead binds candidate object point clouds to persistent video tags, enabling the MLLM to select relevant objects and generate programs over their corresponding 3D variables.

\begin{figure*}[t]
    \centering
    \includegraphics[width=1.\linewidth]{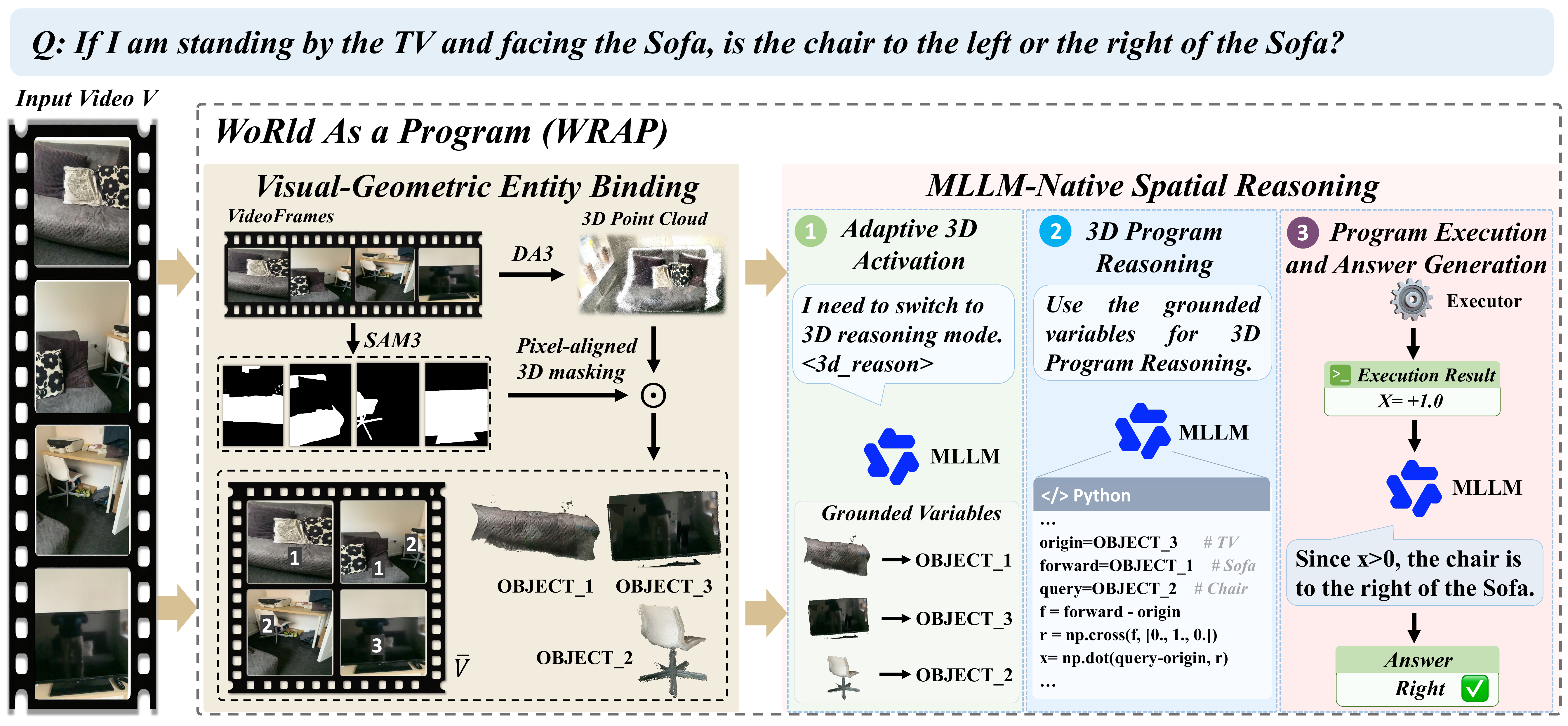}
    \caption{\itbf{Detailed illustration of our proposed \itbf{WRAP} framework.} Given an input video and query, \itbf{WRAP} performs \itbf{Visual-Geometric Entity Binding} to associate visual object instances with their corresponding 3D point clouds as grounded variables (\eg \texttt{OBJECT\_1}). \itbf{WRAP} then performs \itbf{MLLM-Native Spatial Reasoning} in three steps: \ding{202} \itbf{Adaptive 3D Activation} determines whether explicit 3D reasoning is required, \ding{203} \itbf{3D Program Reasoning} formulates an executable geometric program over the grounded variables, and \ding{204} The program is executed to obtain the geometric result, which is then fed back to the same MLLM for final answer generation.}
    \label{fig:framework}
\end{figure*}

\section{WoRld As a Program}
Our \itbf{WRAP} aims to turn visual objects in videos into persistent, grounded 3D entities that MLLMs can directly reference and program over for spatial reasoning. To this end, \itbf{WRAP} binds video objects to addressable 3D entities, termed \textit{Grounded Variables}, and enables the MLLM to formulate executable spatial programs directly over them, while delegating only deterministic computation to an external executor. \itbf{WRAP} consists of two key components: \itbf{Visual-Geometric Entity Binding}, which instantiates 3D objects as grounded variables, and \itbf{MLLM-Native Spatial Reasoning}, which formulates executable spatial programs over these variables (\textit{3D Program Reasoning}). We further introduce a two-stage learning strategy for visual-geometric grounding and programmatic spatial reasoning.

\subsection{Visual-Geometric Entity Binding}
\label{sec:visual_geometric}
Since current MLLMs are primarily designed for video and image inputs, with limited capability to directly process raw 3D point clouds, we first project object instances observed in the video into the reconstructed 3D space to obtain their corresponding point-cloud representations (\textbf{3D Object Instantiation}). We then establish explicit correspondences between the 2D object instances in the video and their instantiated 3D counterparts, enabling 2D visual-to-3D geometric entity binding (\textbf{2D-3D Entity Binding}).

\paragraph{3D Object Instantiation.}
To construct instance-level 3D representations from the input video, we first use an off-the-shelf reconstruction model (\eg \textit{DepthAnything 3~\cite{lin2025depth}}) to estimate per-frame depth, camera intrinsics, and extrinsics. We then prompt the MLLM to identify scene object categories and use them to guide SAM3~\cite{carion2026sam} for instance segmentation. Since SAM3 may fail to preserve instance identity over long videos, especially under long-term disappearance or multiple same-category instances, we further introduce geometry-based instance merging. Each instance mask is back-projected into a shared world coordinate system and aggregated across frames to form a voxelized point cloud $\mathcal{P}_i$. For same category instance pairs, we first filter candidates by 3D bounding-box distance, then measure bidirectional geometric correspondence with KD-trees. Let $||\mathcal{P}_{A}||$ and $||\mathcal{P}_{B}||$ denote the point counts of objects $A$ and $B$. $||\mathcal{P}_{A_m}||$ and $||\mathcal{P}_{B_m}||$ denote the numbers of points with a neighbor in the other point cloud within a predefined radius. We define the geometric overlap score as:
\begin{equation}
    s(A,B)=\max(\frac{||\mathcal{P}_{A_m}||}{||\mathcal{P}_{A}||}, \frac{||\mathcal{P}_{B_m}||}{||\mathcal{P}_{B}||}).
\end{equation}
If $s(A,B) > \alpha$, with $\alpha=0.5$, the two tracks are considered observations of the same physical object and are merged. Pairwise matches are transitively merged using a disjoint-set structure, yielding a unified identity and 3D point cloud for each object.

\paragraph{2D-3D Entity Binding.}
After constructing an instance-level 3D point cloud $\mathcal{P}_i$ for each physical object, we establish explicit correspondences between visual observations and their geometric entities. Rather than feeding raw point clouds into the MLLM, we preserve the native video interface and expose each 3D object through a symbolic visual reference. Inspired by Set-of-Mark~\cite{yang2023set, yuan2026boosting} prompting, we assign each object a unique tag in the video frames $\bar{V}$ and a symbolic identifier $s_i = \texttt{OBJECT}\_i$, referred to as a \textit{Grounded Variable}. This defines a deterministic binding $\Phi: s_i \longleftrightarrow \mathcal{P}_i$, with the same identifier rendered alongside the corresponding visual instance. During 3D program reasoning, any reference to $\texttt{OBJECT}_i$ can therefore be directly resolved to its associated 3D point cloud $\mathcal{P}_i$. This visual-symbolic-geometric binding provides a unified interface between video perception and explicit 3D computation without requiring the MLLM to directly process point-cloud representations.

\subsection{MLLM-Native Spatial Reasoning}
Building on \textbf{Visual-Geometric Entity Binding}, we expose video objects and their associated 3D point clouds as grounded variables to the MLLM. The model performs \textbf{3D Program Reasoning} based on the tagged video $\bar{V}$ and these grounded variables. Since object tags may introduce unnecessary visual clutter for queries that do not require explicit 3D program reasoning (\eg appearance ordering), we adopt \textbf{Adaptive 3D Activation}: the model operates on clean (untagged) video by default and switches to tagged video only when explicit 3D program reasoning is required. Both modes share the same MLLM and differ only in the system prompt and visual input.

\paragraph{3D Program Reasoning.}
Conditioned on the tagged video $\bar{V}$ and query $Q$, the MLLM identifies the required spatial operation and generates an executable program $C$ that explicitly specifies the reasoning procedure, such as relative-direction computation or 3D scene editing. To make grounded variables directly accessible during execution, all grounded variables and their associated 3D representations are registered in the executor \texttt{namespace}, allowing identifiers such as $\texttt{OBJECT}\_i$ to be deterministically resolved to their underlying 3D entities. The generated program can therefore operate directly on scene geometry without additional grounding or task-specific tool selection. Execution is defined as $\text{Exec}(C, E_\Phi, S)\rightarrow (z, S')$, where $E_\Phi$ denotes the executor environment, $S$ and $S'$ are the scene states before and after execution, and $z$ is the execution result. The MLLM then produces the final answer conditioned on the tagged video, query, generated program, execution result, and updated scene state.

\noindent\textbf{Adaptive 3D Activation.}
Given a clean video $V$ and a query $Q$, the MLLM first determines whether explicit 3D program reasoning is necessary. For queries that can be answered directly from the visual input (\eg appearance ordering), the model remains in the standard reasoning mode. When the model emits the designated trigger \texttt{<3d\_reason>}, inference switches to 3D program reasoning, and the clean video $V$ is replaced with its tagged counterpart $\bar{V}$, where each visual object is associated with a grounded variable.

\subsection{Two Stage Training for WRAP}

To equip MLLMs with this reasoning paradigm, we adopt a two-stage learning framework for \itbf{WRAP}. In \textbf{Stage 1}, the MLLM is trained to acquire basic spatial perception capabilities and interpret tagged video frames. In \textbf{Stage 2}, we construct $27$K spatial program learning samples with tagged videos to teach the model when to activate 3D program reasoning and how to reason over grounded variables.

\paragraph{Stage 1: Spatial Grounding.}
In this stage, we equip the MLLM with basic spatial perception capabilities and the ability to interpret tagged video. Given a video $V$, query $Q$, and the corresponding target response $Y=\{y_t\}_{t=1}^T$, we optimize the model using standard autoregressive objective:
\begin{equation}
    \mathcal{L}_{S1}=-\sum_{t=1}^Tlogp_\theta(y_t|y_{<t},V,Q).
\end{equation}
The training data cover diverse spatial perception tasks and include both clean and tagged videos, enabling the model to recognize spatial relationships while learning the semantics of visual tags such as \texttt{OBJECT}\_$i$. This stage provides the perceptual and grounding foundation required for subsequent 3D program reasoning.

\begin{table*}[th]
\caption{\itbf{Comparison of \itbf{WRAP} with existing methods on ReVSI.} Results are reported across numerical and multiple-choice video spatial reasoning tasks. Best results are highlighted in \textbf{bold}.}
\centering
\resizebox{1.\linewidth}{!}{
\begin{tabular}{lccccccccc}
\toprule
\multicolumn{1}{l}{\multirow{2}{*}{\itbf{Method}}}                  & \multicolumn{1}{c}{\multirow{2}{*}{\itbf{Frames}}} & \multicolumn{4}{c}{\itbf{Numerical Question}}         & \multicolumn{3}{c}{\itbf{Multiple-Choice Question}} & \multicolumn{1}{l}{\multirow{2}{*}{\itbf{Avg.}}} \\    \cmidrule(lr){3-6} \cmidrule(lr){7-9}
                       &        & \textit{Obj. Cnt.} & \textit{Abs. Dist.} & \textit{Obj. Size} & \textit{Room Size} & \textit{Rel. Dist.}    & \textit{Rel. Dir.}    & \textit{Route Plan}    &      \\ \hline
\multicolumn{10}{l}{\itbf{Proprietary Models}}                                                                                                \\ \hdashline
GPT-5.2                & 64     & 56.2      & 41.5       & 73.9      & \textbf{63.0}      & 48.4          & 34.9         & 38.2          & 50.9 \\
Gemini 3 Flash         & 1 FPS  & \textbf{65.7}      & 53.1       & 77.6      & 52.8      & 64.6          & 47.9         & 41.8          & 57.6 \\
Gemini 3 Pro           & 1FPS   & 60.1      & 54.7       & \textbf{79.3}      & 51.9      & 68.1          & 56.0         & 56.4          & 60.9 \\ \hline
\multicolumn{10}{l}{\itbf{Open-Source General Models}}                                                                                        \\ \hdashline
LLaVA-Video-72B-Qwen2~\cite{zhang2024llava}  & 64     & 40.1      & 29.6       & 59.3      & 27.9      & 39.6          & 24.8         & 43.0          & 37.8 \\
Qwen3-VL-8B-Instruct~\cite{bai2025qwen3}   & 64     & 40.4      & 52.3       & 69.0      & 45.1      & 57.1          & 39.5         & 40.5          & 49.1 \\
Qwen3-VL-32B-Instruct~\cite{bai2025qwen3}  & 64     & 46.9      & 65.0       & 70.4      & 55.8      & 53.8          & 34.0         & 47.3          & 53.3 \\
InternVL3.5-38B~\cite{wang2025internvl3}        & 64     & 43.8      & 60.6       & 70.2      & 58.4      & 57.4          & 45.9         & 42.7          & 54.1 \\ \hline
\multicolumn{10}{l}{\itbf{Spatially Specialized Models}}                                                                                      \\ \hdashline
SpaceR-7B~\cite{ouyang2025spacer}         & 32     & 30.7      & 34.5       & 52.0      & 18.6      & 22.8          & 34.5         & 20.2          & 30.5 \\
Qwen2.5-VL-7B-Instruct & 4 FPS  & 36.9      & 15.0       & 49.7      & 29.0      & 31.5          & 29.5         & 36.7          & 32.6 \\
Spatial-MLLM-4B-820K~\cite{wu2026spatial}   & 16     & 41.5      & 40.0       & 53.1      & -         & 30.7          & 39.2         & -             & 40.9 \\
VST-7B-SFT~\cite{yang2026visual}            & 4 FPS  & 35.4      & 52.6       & 67.9      & 47.2      & 49.2          & 36.9         & 35.4          & 46.4 \\
Cambrian-S-7B~\cite{yang2026cambrian}          & 128    & 48.4      & 60.5       & 65.5      & 46.7      & 37.1          & 48.5         & 37.0          & 49.1 \\
LLaVA-Video-7B-Qwen2   & 32     & 29.9      & 1.5        & 53.0      & 19.3      & 39.1          & 33.8         & 38.8          & 30.8 \\
VLM3R-7B~\cite{fan2026vlm}               & 32     & 41.6      & 61.6       & 64.8      & 52.5      & 46.5          & 49.5         & 34.1          & 50.1 \\
GeoThinker 7B~\cite{li2026thinking}   & 64     &   38.6    &  67.0    &  64.2     &   54.7    &     40.5      &    46.8      &    41.2       & 50.4 \\
GeoThinker 8B~\cite{li2026thinking} & 64     &  37.6     &   73.7   &   66.8    &   55.0    &    62.5       &     48.5     &     41.6      & 55.1 \\
S-Agent~\cite{dai2026s}                & 64     & 54.0      & 45.6       & 62.6      & 53.4      & 63.6          & 66.4         & 66.1          & 58.8 \\
 \tablerowcolor \itbf{WRAP} 2B     & 64     &  35.5     &   75.9     &   62.7    &  45.2     &    68.7       &    84.1      &     60.9      & 61.9  \\
 \tablerowcolor \itbf{WRAP} 4B     & 64     &   40.1    &   75.8     &   63.9    &  53.1     &     73.4      &     81.7     &    58.8       & 63.8  \\
 \tablerowcolor \itbf{WRAP} 9B     & 64     &  40.2    &   \textbf{77.1}     &   64.3    &  51.6     &  \textbf{74.7}         &     \textbf{86.6}     &    \textbf{68.5}       & \textbf{66.1} \\
\bottomrule
\end{tabular}
}
\label{table:revsi}
\end{table*}

\paragraph{Stage 2: Spatial Program Learning.}
In this stage, we further train the MLLM to determine when explicit 3D program reasoning is required and to generate executable programs over grounded variables. The objective consists of three terms: $\mathcal{L}_{adaptive}$ supervises the reasoning-mode decision, encouraging the model to either remain in the standard reasoning mode or emit the \texttt{<3d\_reason>} trigger. $\mathcal{L}_{3D}$ supervises program generation and the corresponding final answer under tagged video input. $\mathcal{L}_{standard}$ supervises direct answering from the clean video. Formally, we optimize
\begin{equation}
\mathcal{L}_{S2}=\mathcal{L}_{\text{adaptive}}+\mathcal{L}_{\text{3D}}+\mathcal{L}_{\text{standard}}.
\end{equation}
Here, $\mathcal{L}_{adaptive}=-logp_\theta(r|V,Q)$ supervises the target reasoning mode, where $r \in \{\text{standard}, \text{3D}\}$ denotes the target reasoning mode, with the 3D mode corresponding to emission of the \texttt{<3d\_reason>} trigger. $\mathcal{L}_{3D}$ and $\mathcal{L}_{standard}$ are autogressive cross-entropy losses over the 3D program reasoning and standard reasoning trajectories, conditioned on $(\bar{V}, Q)$ and $(V, Q)$, respectively.

\section{Experiments}

\subsection{Experimental Setup}
\label{sec:experimental_setup}

\paragraph{Benchmarks.}
We evaluate \itbf{WRAP} on four benchmarks spanning video spatial reasoning tasks and 3D scene editing tasks. For video spatial reasoning, VSI-Bench~\cite{yang2025thinking} covers metric and spatiotemporal reasoning from egocentric videos, while VSI-Bench-Debiased~\cite{ii2026benchmark} reduces non-visual shortcuts to better isolate visually grounded reasoning. ReVSI~\cite{zhang2026revsi} further improves evaluation reliability by correcting annotation errors and ensuring answerability under the given frame budget. For 3D scene editing, EditRoom-DB~\cite{zheng2025editroom} evaluates language-guided room-layout manipulation, including object translation, rotation, scaling, replacement, addition, and removal.

\paragraph{Models.}
We adopt Qwen3.5~\cite{qwen35blog} with 2B, 4B, and 9B parameters as the MLLM backbones. SAM3~\cite{carion2026sam} is used to extract object-level segmentation masks from the input videos, while DepthAnything 3~\cite{lin2025depth} reconstructs the corresponding 3D point clouds. Specifically, we employ DA3NESTED-GIANT-LARGE 1.1 for metric 3D reconstruction, enabling geometric measurements in real-world scale. We additionally extract the floor point cloud to support spatial reasoning relative to the scene geometry.

\begin{table*}[th]
\caption{\itbf{Comparison of \itbf{WRAP} with existing methods on VSI-Bench.} Results cover numerical and multiple-choice video spatial reasoning tasks, with the best results in bold.}
\centering
\resizebox{1.\linewidth}{!}{
\begin{tabular}{lccccccccc}
\toprule
\multicolumn{1}{l}{\multirow{2}{*}{\itbf{Method}}}                  & \multicolumn{4}{c}{\itbf{Numerical Question}}         & \multicolumn{4}{c}{\itbf{Multiple-Choice Question}} & \multicolumn{1}{l}{\multirow{2}{*}{\itbf{Avg.}}} \\    \cmidrule(lr){2-5} \cmidrule(lr){6-9}
                           & \textit{Obj. Cnt.} & \textit{Abs. Dist.} & \textit{Obj. Size} & \textit{Room Size} & \textit{Rel. Dist.}    & \textit{Rel. Dir.}    & \textit{Route Plan} & \textit{Appr. Order}    &      \\ \hline
\multicolumn{10}{l}{\itbf{Baseline}}                                                                                                          \\ \hdashline
Chance (Random)         &  -        &  -         &  -        & -         & 25.0         & 36.1         & 28.3    &  25.0    & -    \\
Chance (Frequency)       & 62.1      & 32.0       & 29.9      & 33.1      & 25.1          & 47.9        & 28.4       & 25.2 & 34.0 \\ \hline
\multicolumn{10}{l}{\itbf{General-purpose Models}}                                                                                                \\ \hdashline
Qwen2.5VL 7B~\cite{Qwen2.5-VL} &   25.2    &   10.5     &   36.4    &  29.6     &       38.4    &   38.0       &    29.8       & 26.8 &  29.3 \\
GPT-4o~\cite{hurst2024gpt}                  & 46.2      & 5.3       & 43.8      & 38.2      & 37.0          & 41.3         & 31.5          & 28.5 & 34.0  \\
InternVL-3 8B~\cite{zhu2025internvl3}  &    68.1   &  39.0      &   48.4    &  33.6     &      48.3     &   36.4       &    27.3       &  35.4 & 42.1  \\
Gemini-2.5 Pro~\cite{comanici2025gemini}    &  43.8     &   34.9     &   64.3    &  42.8     &   61.1        &    47.8      &      45.9     & 71.3 & 51.5  \\ \hline
\multicolumn{9}{l}{\itbf{Spatially Specialized Models}}                                                                                      \\ \hdashline
VG-LLM 8B~\cite{zheng2026learning}  &   67.9    &   37.7     &  58.6     &    62.0   &  46.6         &    40.7      &   32.4        & 59.2 &  50.7 \\
Cambrian-S 3B~\cite{yang2026cambrian}  &   70.7    &  40.6      &  68.0     &   46.3    &     64.8      &    61.9      &    27.3       & 78.8 & 57.3  \\
3DThinker 3B~\cite{chen2026think} &   -    &   -     &  -     &   -    &    -       &    -      &     -      & - & 59.1  \\
VLM-3R 7B~\cite{fan2026vlm}  &  70.2     &    49.4    &   69.2    &   67.1    &   65.4        &    80.5      &     45.4      & 40.1 & 60.9  \\
VST 7B~\cite{yang2026visual}  &   71.6    &   43.8     &  75.5     &   69.2    &    60.0       &    55.6      &     44.3      & 69.2 & 61.2  \\
3DThinker 7B~\cite{chen2026think} &   -    &   -     &  -     &   -    &    -       &    -      &     -      & - & 63.7  \\
Cambrian-S 7B~\cite{yang2026cambrian}            & \textbf{73.2}      & 50.5       & \textbf{74.9}     & \textbf{72.2}      & \textbf{71.1}          & 76.2         & 41.8          & \textbf{80.1} & 67.5  \\
GeoThinker 7B~\cite{li2026thinking} &   -    &   -     &  -    &   -    &   -        & -         &     -      & - & 68.5  \\
GeoThinker 8B~\cite{li2026thinking}  &    -   &  -      &   -   &   -    &   -        &       -   &     -      & - & \textbf{72.6}  \\
\tablerowcolor \itbf{WRAP 2B}       &   69.1    &    68.1    &   71.6    &   65.2    &     65.2      &      85.8    &      47.9     & 76.9 & 68.7  \\
\tablerowcolor \itbf{WRAP 4B}       &   68.8    &   \textbf{69.2}     &   74.2    &   67.1    &   67.5        &    \textbf{87.5}      &     \textbf{50.0}      & 73.3 & 69.7 \\
\tablerowcolor \itbf{WRAP 9B}       &   68.5   &   68.9     &   74.5    &   66.9    &   67.6        &     87.2     &    45.4       &  76.7 & 69.5 \\
\bottomrule
\end{tabular}
}
\label{table:vsibench}
\end{table*}

\paragraph{Training Data.}
For Stage 1, we train on approximately 700K samples from SoM-LLaVA~\cite{yan2024list} and VSI-590K~\cite{yang2026cambrian} to establish robust spatial grounding and general spatial perception. For Stage 2, we construct 25K program-oriented spatial reasoning samples from ScanNet~\cite{dai2017scannet} and incorporate an additional 2K EditRoom-DB~\cite{zheng2025editroom} examples to extend the learned reasoning paradigm to 3D scene editing.

\subsection{Results on Video Spatial Reasoning}

\paragraph{Results on ReVSI.}
As shown in Table~\ref{table:revsi}, our \itbf{WRAP} consistently outperforms both general-purpose MLLMs and spatially specialized baselines, demonstrating the benefit of programmable 3D reasoning. Among the evaluated variants, \itbf{WRAP} 9B achieves the highest average score among the compared methods on ReVSI, reaching 66.1\%. This exceeds Gemini 3 Pro and S-Agent~\cite{dai2026s} by 5.2 and 7.3 points, respectively. \itbf{WRAP} 9B also obtains scores of 77.1 on absolute distance, 74.7 on relative distance, and 86.6 on relative direction. The results show strong performance on these geometry-intensive tasks.

\paragraph{Results on VSI-Bench.}
As shown in Table~\ref{table:vsibench}, our \itbf{WRAP} achieves strong performance across model scales, with \itbf{WRAP} 4B reaching an average score of 69.7\% and showing particularly strong results on geometry-intensive tasks such as absolute distance and relative direction. Notably, even \itbf{WRAP} 2B attains an average score of 68.7\%, demonstrating that programmable 3D representations can substantially enhance spatial reasoning with a compact backbone. Although \itbf{WRAP} does not surpass GeoThinker 8B~\cite{li2026thinking}, GeoThinker is trained on approximately 1.8M samples, compared with about 727K for \itbf{WRAP}, while \itbf{WRAP} remains competitive on several geometry-intensive tasks.

\begin{table*}[t]
\caption{\itbf{Comparison of \itbf{WRAP} with existing methods on VSI-Bench-Debiased.} The best results are highlighted in bold.}
\centering
\resizebox{1.\linewidth}{!}{
\begin{tabular}{lccccccccc}
\toprule
\multicolumn{1}{l}{\multirow{2}{*}{\itbf{Method}}}                  & \multicolumn{4}{c}{\itbf{Numerical Question}}         & \multicolumn{4}{c}{\itbf{Multiple-Choice Question}} & \multicolumn{1}{l}{\multirow{2}{*}{\itbf{Avg.}}} \\    \cmidrule(lr){2-5} \cmidrule(lr){6-9}
                           & \textit{Obj. Cnt.} & \textit{Abs. Dist.} & \textit{Obj. Size} & \textit{Room Size} & \textit{Rel. Dist.}    & \textit{Rel. Dir.}    & \textit{Route Plan} & \textit{Appr. Order}    &      \\ \hline
\multicolumn{10}{l}{\itbf{General-purpose Models}}                                                                                                \\ \hdashline
LLaVA-Video-7B~\cite{zhang2024llava}  &   38.6    &   15.4   &   34.7   &   22.1    &   42.6   &  37.9     &   20.2   & 34.0 & 30.7  \\
LLaVA-Video-72B~\cite{zhang2024llava}     &   37.3    &  22.3    &  45.8    &   33.7    &   43.9   &    37.8   &  25.4    &  48.1 & 36.8  \\ 
LLaVA-OneVision-7B~\cite{li2024llava}      &   37.1    &   16.3   &  31.4    &  18.0     &   39.7   &   34.2    &   23.7   &  27.7 &  28.5 \\ 
LLaVA-OneVision-72B~\cite{li2024llava}     &   32.0    &  21.8    &   44.3   &  40.8     &   39.0   &   37.3    &   21.9   &  47.8 & 35.6  \\ 
Gemini-2.5 Pro~\cite{comanici2025gemini}     &   40.8    &  34.4    &   56.7   &   40.9    &  61.0    &   46.8    &    43.0  & 69.5 & 49.1  \\ 
Qwen3.5 2B~\cite{qwen35blog} &    36.3   &    21.3    &    31.5   &   29.7    &   44.5        &    38.6      &      24.6     & 46.5 & 34.1  \\
Qwen3.5 4B~\cite{qwen35blog} &    48.2   &   32.4     &    57.2   &  50.8     &     57.1      &   46.4       & 28.1          & 68.2 & 48.6  \\
Qwen3.5 9B~\cite{qwen35blog} &    49.5   &  33.7      &  59.6     &  63.0     &  67.4         &  58.0        &  34.2         & 72.3 & 54.7  \\ \hline
\multicolumn{9}{l}{\itbf{Spatially Specialized Models}}                                                                                      \\ \hdashline
Qwen3.6-27B w/ SpatialClaw~\cite{cho2026spatialclaw} &   -    &   -     &  -     &   -    &    -       &    -      &     -      & - &  57.9 \\
Cambrian-S-7B~\cite{yang2026cambrian} &   -    &   -     &  -     &   -    &    -       &    -      &     -      & - &  59.9 \\
VG-LLM 8B~\cite{zheng2026learning} &   -    &   -     &  -     &   -    &    -       &    -      &     -      & - &  55.2 \\
GeoThinker 3B~\cite{li2026thinking}       &   41.4    &   26.9    &  42.2    &  52.5    &     50.6     &     38.5    &    21.1       & 73.3 & 43.3  \\
GeoThinker 7B~\cite{li2026thinking}       &  \textbf{60.4}     &   53.3    &  60.7    &  73.9    &   68.7   &   79.2      &   40.4        & 62.6 &  62.4 \\
GeoThinker 8B~\cite{li2026thinking}       &    59.8   &  59.7     &  \textbf{64.6}    &   \textbf{75.6}   &      \textbf{71.0}    &   81.2      &    \textbf{48.2}       & \textbf{72.3} & \textbf{66.6}  \\
\tablerowcolor \itbf{WRAP 2B}       &   53.1    &     64.3  &   57.2   &  65.3    &    63.2      &  84.8       &     34.2      & 76.4 & 62.3  \\
\tablerowcolor \itbf{WRAP 4B}       &  59.5     & 66.8       &    60.2   &   67.9    &    67.4      &      82.4    &   33.3       & 71.1 & 63.6 \\
\tablerowcolor \itbf{WRAP 9B}       &   55.8   &   \textbf{67.1}     &   59.9    &   66.5    &   65.2        &     \textbf{84.8}     &    35.1       &  70.4 & 63.1 \\
\bottomrule
\end{tabular}
}
\label{table:vsibench_debiased}
\end{table*}

\paragraph{Results on VSI-Bench-Debiased.}
Table~\ref{table:vsibench_debiased} compares \itbf{WRAP} with general-purpose and spatially specialized models. \itbf{WRAP} 9B achieves an average score of \textbf{63.1}, outperforming Gemini-2.5 Pro, Cambrian-S-7B~\cite{yang2026cambrian}, and VG-LLM 8B by 14.0\%, 3.2\%, and 7.9\% points, respectively. It also achieves the best performance on \textit{Absolute Distance} (\textbf{67.1\%}) and \textit{Relative Direction} (\textbf{84.8\%}), demonstrating strong metric and directional reasoning capability.

\begin{table*}[t]
\centering
\caption{\itbf{Comparison of \itbf{WRAP} with prior methods on EditRoom-DB} across bedroom, dining room, and living room scenes using IoU, S-IoU, LPIPS, and CLIP metrics. \textit{No-Op} is a baseline that leaves the input scene unchanged, measuring the score achievable without executing the requested edit. $\dagger$ denotes results reproduced by us using the official implementation.}
\resizebox{1.\linewidth}{!}{
\begin{tabular}{lcccccccccccc}
\toprule
\multirow{2}{*}{\itbf{Method}} & \multicolumn{4}{c}{\itbf{Bedroom}}                                                               & \multicolumn{4}{c}{\itbf{Dining room}}                                                           & \multicolumn{4}{c}{\itbf{Living room}}                                                           \\ \cmidrule(lr){2-5} \cmidrule(lr){6-9} \cmidrule(lr){10-13} 
                        & \textit{IOU} $\uparrow$                  & \textit{S-IOU} $\uparrow$              & \textit{LPIPS} $\downarrow$               & \textit{CLIP} $\uparrow$                 & \textit{IOU} $\uparrow$                 & \textit{S-IOU} $\uparrow$               & \textit{LPIPS} $\downarrow$               & \textit{CLIP} $\uparrow$                & \textit{IOU} $\uparrow$                 & \textit{S-IOU} $\uparrow$               & \textit{LPIPS} $\downarrow$               & \textit{CLIP} $\uparrow$                \\ \hline
\textit{No-Op}   &    0.8539           &    0.8506            &       0.1007         &    0.9651           &       0.9257        &     0.9240           &   0.0492            &        0.9807       &     0.9287          &          0.9269     &    0.0470        &      0.9836         \\ \hdashline
DiffuScene-N~\cite{tang2024diffuscene}            & 0.6220               & 0.6125               & 0.1431               & 0.9415               & 0.4473               & 0.4321               & 0.1921               & 0.9254               & 0.4332               & 0.4161               & 0.1810               & 0.9236               \\
SceneEditor-N~\cite{zheng2025editroom}           & 0.7055               & 0.6942               & 0.1261               & 0.9510               & 0.5262               & 0.5091               & 0.1557               & 0.9366               & 0.4618               & 0.4498               & 0.1738               & 0.9329               \\
EditRoom~\cite{zheng2025editroom}                & 0.7342               & 0.7236               & 0.1090               & 0.9597               & 0.5360               & 0.5196               & 0.1460               & 0.9460               & 0.4710               & 0.4629               & 0.1616               & 0.9446               \\
EditRoom$^\dagger$~\cite{zheng2025editroom} &    0.7278            &     0.7063           &     0.1326          &     0.9503           &   0.5402             &     0.5193           &     0.1529           &      0.9416          &       0.4699         &      0.4136          &    0.1726            &     0.9369           \\
\tablerowcolor \itbf{WRAP} 2B               & \textbf{0.9224} & \textbf{0.9221} & 0.0432  & \textbf{0.9853} & \textbf{0.9333} & \textbf{0.9330} & 0.0323 & \textbf{0.9885} & \textbf{0.9438} & \textbf{0.9434} & \textbf{0.0265} & \textbf{0.9895} \\
\tablerowcolor \itbf{WRAP} 4B               & 0.9215 & 0.9212 & \textbf{0.0417} & 0.9852 & 0.9293 & 0.9289 & 0.0338 & 0.9866 & 0.9360 & 0.9356 & 0.0287 & 0.9845 \\ 
\tablerowcolor \itbf{WRAP} 9B & 0.9143 & 0.9140 & 0.0445 & 0.9811 & 0.9324 & 0.9320 & \textbf{0.0311} & 0.9861 & 0.9375 & 0.9371 & 0.0283 & 0.9886 \\ 
\bottomrule
\end{tabular}
}
\label{table:editroom}
\end{table*}

\subsection{Results on EditRoom-DB}
\label{appendix:editroom}
Table~\ref{table:editroom} presents the quantitative comparison on EditRoom-DB across three scene categories. \itbf{WRAP} consistently outperforms existing methods across four metrics, demonstrating superior fidelity, perceptual similarity, and semantic alignment. Additionally, we introduce Edit-IoU and Edit-S-IoU, which evaluate geometric consistency specifically on th objects targeted by the editing instructions. As shown in Table~\ref{table:editroom_edit}, \itbf{WRAP} consistently outperforms EditRoom and the No-Op baseline across all scene categories, demonstrating its superior ability to accurately execute object-level edits while preserving geometric consistency. These improvements remain consistent across different model scales, further validating the effectiveness and robustness of \itbf{WRAP}.

\begin{table*}[h]
\centering
\caption{\itbf{Object-focused evaluation on EditRoom-DB} using the proposed Edit-IoU and Edit-S-IoU metrics across bedroom, dining room, and living room scenes. \textit{No-Op} denotes leaving the original scene unchanged without performing the requested edit.  $\dagger$ denotes results from our reimplementation using the official codebase.}
\begin{tabular}{lcccccc}
\toprule
    \multirow{2}{*}{\itbf{Method}}     & \multicolumn{2}{c}{\itbf{Bed Room}} & \multicolumn{2}{c}{\itbf{Dining Room}} & \multicolumn{2}{c}{\itbf{Living Room}} \\
    \cmidrule(lr){2-3} \cmidrule(lr){4-5} \cmidrule(lr){6-7}     & \textit{EDIT-IOU}$\uparrow$     & \textit{EDIT-S-IOU}$\uparrow$    & \textit{EDIT-IOU}$\uparrow$      & \textit{EDIT-S-IOU}$\uparrow$      & \textit{EDIT-IOU}$\uparrow$      & \textit{EDIT-S-IOU}$\uparrow$      \\ \hline
\textit{No-Op} &   0.2377    &    0.2202     &     0.2120          &         0.1939        &     0.2242    &   0.2047        \\ \hdashline
EditRoom$^\dagger$~\cite{zheng2025editroom} & 0.4594       & 0.4344        &   0.2456            &      0.2233           & 0.2659        & 0.2233          \\
\tablerowcolor \itbf{WRAP 2B}  & \textbf{0.6784}       & \textbf{0.6774}        & \textbf{0.5589}        & \textbf{0.5568}          & \textbf{0.5708}        & \textbf{0.5688}          \\
\tablerowcolor \itbf{WRAP 4B}  & 0.6683       & 0.6673        & 0.5211        & 0.5189          & 0.5555        & 0.5535          \\
\tablerowcolor \itbf{WRAP 9B}  & 0.6654       & 0.6645        & 0.5503        & 0.5482          & 0.5509        & 0.5489     \\
\bottomrule
\end{tabular}
\label{table:editroom_edit}
\end{table*}

\subsection{Ablation Study}
To better understand the contribution of each component in \itbf{WRAP}, we conduct ablation studies with the 4B model on ReVSI. As shown in Table~\ref{table:ablation}, the baseline (first line) uses the same training data described in Section~\ref{sec:experimental_setup} and is constrained to reason directly from the original video frames at inference time. Introducing Tagged Video improves ReVSI from $45.5\%$ to $50.8\%$. This suggests that object tags facilitate explicit object referencing. Incorporating \itbf{3D Program Reasoning} further improves performance to $61.9\%$ on ReVSI, highlighting the benefit of executable reasoning over grounded variables. 
Finally, \itbf{Adaptive 3D Activation} achieves the best performance on the benchmark, reaching $63.8\%$, indicating the benefit of selectively invoking tagged videos and explicit 3D reasoning.

\begin{table}[hbpt]
\centering
\caption{\itbf{Ablation study of the key components in \itbf{WRAP} 4B on ReVSI.} ``\itbf{Tagged}'' represents the Tagged Video, ``\itbf{Adaptive}'' represents the Adaptive 3D Activation.}
\begin{tabular}{ccc|c}
\toprule
  \itbf{Tagged}  & \itbf{3D Program Reasoning} & \itbf{Adaptive}  & \itbf{ReVSI} \\ \hline
                             &           &   &  45.5        \\
       \checkmark    &        &                      &   50.8            \\
       \checkmark & \checkmark &  & 61.9  \\
       \checkmark    &   \checkmark       & \checkmark    &    \textbf{63.8}        \\
 \bottomrule
\end{tabular}
\label{table:ablation}
\end{table}

\subsection{Comparison with Implicit Spatial Modeling method}
\label{appendix:consistent_result}
To investigate whether implicit spatial modeling methods lack stable and consistent spatial representations, we evaluate prediction consistency ob the absolute-distance task, where the ground-truth distances are invariant to video frame order. Specifically, we feed each model the original and temporally reversed video frames and measure the agreement between their predictions using the Concordance Correlation Coefficient (CCC). As shown in Table~\ref{table:consistency}, \itbf{WRAP} achieves a substantially higher CCC (94.0) than implicit spatial modeling methods, indicating more consistent spatial reasoning. Moreover, \itbf{WRAP} maintains its accuracy under temporal reversal (58.0\% vs. 57.3\%), while competing methods exhibit larger performance drops, demonstrating the robustness of our explicit spatial modeling.

\begin{table}[h]
\centering
\caption{\itbf{Comparison with implicit spatial modeling methods.} ``\itbf{CCC}'' represents the Concordance Correlation Coefficient which measures prediction consistency between original and temporally reversed video frames. \itbf{Accuracy} and \itbf{Accuracy$_r$} denote accuracy on the original and reversed frames, respectively.}
\begin{tabular}{lccc}
\toprule
\itbf{Method}            & \itbf{CCC}  & \itbf{Accuracy} & \itbf{Accuracy$_{\text{r}}$} \\ \hline
Qwen3.5 4B~\cite{qwen35blog}    & 82.0 & 33.8 & 30.2   \\
GeoThinker 3B~\cite{li2026thinking} & 84.0 & 35.0 &  32.3  \\
\itbf{WRAP 4B}       & \textbf{94.0} & \textbf{58.0}  & \textbf{57.3} \\ \bottomrule
\end{tabular}
\label{table:consistency}
\end{table}

\subsection{Discussion}
In this section, we discuss the factors underlying the spatial reasoning performance of \itbf{WRAP}, focusing on object grounding and program generation. Specifically, we examine the accuracy of each process, their joint impact on reasoning, and the quality of object-level 3D representations compared with existing explicit-geometry methods. We also assess the flexibility of program generation on unseen spatial reasoning tasks. To support the analysis of grounding and program correctness, we follow ReVSI~\cite{zhang2026revsi} to construct a 1K-sample test set and annotate it with oracle object groundings and programs derived from ScanNet~\cite{dai2017scannet}. The discussion is organized around the following questions:

\begin{figure}[h]
    \centering
    \includegraphics[width=1.\linewidth]{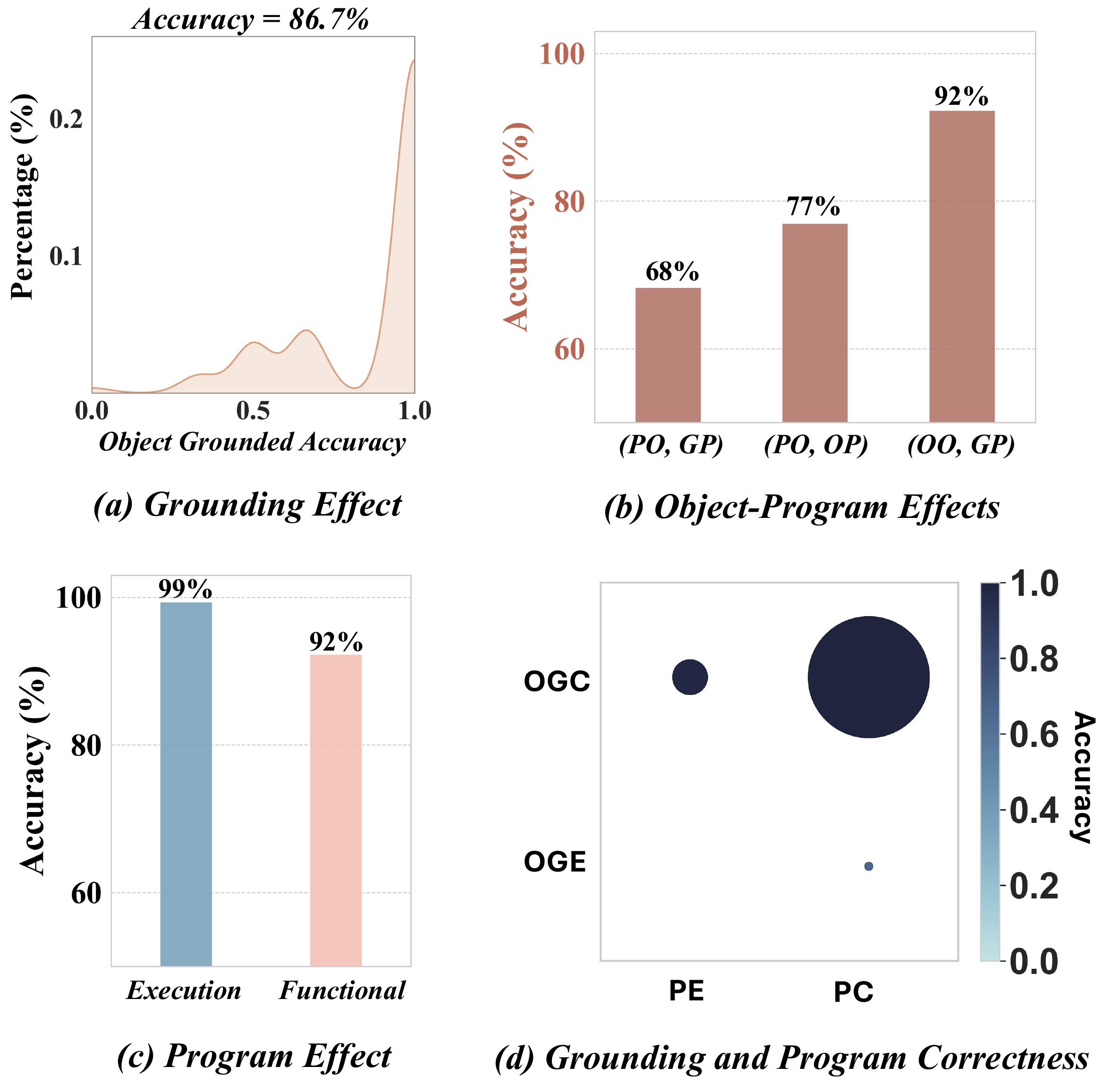}
    \caption{\itbf{Mechanistic analysis of Object Grounding and Program Reasoning in \itbf{WRAP} 4B.} The ``\itbf{PO}'' denotes the predicted grounded object, the ``\itbf{GP}'' denotes the generated program, the ``\itbf{OP}'' denotes the oracle program, and the ``\itbf{OO}'' denotes the oracle grounded object.}
    \label{fig:why_work}
\end{figure}

\paragraph{\itbf{(1) Can \textit{WRAP} Accurately Ground Objects to Grounded Variables?}}
To assess the object grounding accuracy of \itbf{WRAP} during spatial reasoning, we extract the grounded variables (\ie \texttt{OBJECT\_X}) generated during reasoning together with their semantic object names (\eg TV and Bed), and compare the predicted grounded objects against the oracle grounded objects. As shown in Figure~\ref{fig:why_work} (a), \itbf{WRAP} achieves an average grounding accuracy of 86.7\% on the 1K-sample diagnostic set, with most samples concentrated around 90\%. To further isolate the effect of grounding quality, we keep the generated programs fixed and replace the predicted grounded objects with oracle grounded objects, \ie $(PO,GP)\rightarrow(OO,GP)$. As shown in Figure~\ref{fig:why_work} (b), this substitution improves spatial reasoning accuracy from $68\%$ to $92\%$. These results demonstrate that \itbf{reliable visual-geometric grounding is a key prerequisite for accurate spatial reasoning and provides a strong foundation for subsequent 3D program reasoning.}

\paragraph{\itbf{(2) Does WRAP Produce More Accurate Object-Level 3D Representations?}} Beyond selecting the correct objects, spatial reasoning also depends on the quality of their reconstructed geometry. We compare \itbf{WRAP} with strategies adapted from VADAR~\cite{marsili2025visual} and SpatialClaw~\cite{cho2026spatialclaw}. All methods use Qwen3.5-27B without additional training, SAM3 for segmentation, and DepthAnything 3 for 3D reconstruction. The VADAR-based strategy uses GroundingDINO~\cite{liu2023grounding} for object localization, whereas the SpatialClaw-based strategy relies on MLLM-predicted coordinates. \itbf{WRAP} follows the approach described in Section~\ref{sec:visual_geometric}. As shown in Table~\ref{table:compare_existed_grounding}, \itbf{WRAP} achieves a 3D IoU of 38.9\%, compared with 32.5\% and 32.9\% for SpatialClaw and VADAR, respectively, alongside higher downstream task accuracy. The results highlight the complementary roles of accurate object geometry and target identification: \itbf{the former supports reliable geometric computation, while the latter ensures that the computation operates on the intended objects.}
By constructing object geometry before query-time grounding by the MLLM, \itbf{WRAP} allows the object representations to be refined independently of target selection. This design provides a stronger geometric basis for subsequent spatial reasoning.

\begin{table}[t]
\centering
\caption{\itbf{Comparison with existing geometry-based methods.} ``\itbf{3D-IOU}'' evaluate 3D reconstruction quality. ``\itbf{ID Acc.}'' measures object grounding accuracy, while ``\itbf{Task Acc.}'' measures downstream task performance.}
\begin{tabular}{lccc}
\toprule
\itbf{Method}        & \itbf{3D-IoU} & \itbf{ID Acc.} & \itbf{Task Acc.} \\ \hline
SpatialClaw-based  & 32.5   & 52.7    & 25.9 \\
VADAR-based        & 32.9   & 54.4    & 27.9 \\
\itbf{WRAP}         & \textbf{38.9}   & \textbf{67.8}    & \textbf{36.2} \\
\bottomrule
\end{tabular}
\label{table:compare_existed_grounding}
\end{table}

\paragraph{\itbf{(3) Can WRAP Generate Correct 3D Reasoning Programs?}}
To assess the quality of 3D reasoning programs generated by \itbf{WRAP}, we evaluate them from two complementary perspectives: \itbf{executability} and \itbf{functional correctness}. Executability measures whether a program is syntactically and operationally complete enough to run successfully, which we quantify as the fraction of programs that terminate without execution errors. The functional correctness further requires it to produce the same output as the oracle program given the same oracle grounded objects. As shown in Figure~\ref{fig:why_work} (c), the generated program achieve a $99\%$ execution success rate and $92\%$ functional correctness on the diagnostic set. Moreover, replacing the generated programs with oracle programs while keeping the predicted grounded objects fixed improves spatial reasoning accuracy from $68\%$ to $77\%$. These results show that \itbf{WRAP} generates \itbf{highly executable and functionally reliable 3D reasoning programs}, while further improving program generation can still yield \itbf{stronger spatial reasoning performance.}

\paragraph{\itbf{(4) How Do Grounding and 3D Program Reasoning Jointly Affect Spatial Reasoning Performance?}}
To examine the joint effect of object grounding and program correctness on final spatial reasoning performance, we categorize each sample into four combinations based on whether object grounding is correct or erroneous (\textbf{OGC}/\textbf{OGE}) and whether the generated program is correct or erroneous (\textbf{PC}/\textbf{PE}). As shown in Figure~\ref{fig:why_work} (d), the bubble size denotes the number of samples, while color intensity indicates answer accuracy. The results show that high spatial reasoning accuracy is achieved primarily when \itbf{both object grounding and program generation are correct}.

\begin{table}[h]
\centering
\caption{\itbf{Generalization performance (accuracy, \%) on unseen spatial reasoning tasks.} Both methods are trained exclusively on closest-point-based absolute distance estimation.}
\begin{tabular}{lcc}
\toprule
\itbf{Method}       & \itbf{Rel. Dist.} & \itbf{Centroid-based Abs.} \\ \hline
1) Fixed Tool & 65.7      & 22.1                \\
2) Our \itbf{WRAP}   & \textbf{67.9}      & \textbf{55.3}   \\
\bottomrule
\end{tabular}
\label{table:unseen_task}
\end{table}

\paragraph{\itbf{(5) How Well Does WRAP Generalize to Unseen Tasks?}} 
To evaluate generalization to unseen tasks, we fine-tune Qwen3.5-9B with LoRA~\cite{hu2021lora} on 1K closest-point-based absolute distance samples, using either tool calling or \itbf{WRAP}. We evaluate both approaches on unseen relative distance and centroid-based absolute distance tasks, where centroids are defined as the mean of point cloud coordinates. As shown in Table~\ref{table:unseen_task}, both methods achieve comparable accuracy on relative distance estimation. However, on centroid-based distance estimation, \itbf{WRAP} substantially outperforms fixed tool (55.3\% vs. 22.1\% accuracy) by dynamically generating code to compute centroid distances rather than relying on predefined tools. These results demonstrate the \itbf{generalization of \itbf{WRAP} to unseen spatial reasoning tasks.}

\section{Conclusion}
In this paper, we propose \itbf{WoRld As a Program (WRAP)}, a novel framework that transforms reconstructed 3D scenes into addressable and programmable reasoning spaces for MLLMs, enabling more reliable spatial reasoning. Extensive experiments on video spatial reasoning and 3D scene editing benchmarks demonstrate the effectiveness of \itbf{WRAP}. Our analyses lead to three main findings: (1) Accurate visual-geometric grounding provides a strong foundation for subsequent 3D program reasoning. (2) Higher-quality 3D reasoning programs are consistently associated with stronger downstream spatial reasoning performance. (3) Final spatial reasoning performance is closely linked to the joint quality of visual-geometric grounding and 3D program reasoning. In future work, we aim to extend \itbf{WRAP} toward richer 4D world representations.

{
    \small
    \bibliographystyle{ieeenat_fullname}
    \bibliography{main}
}

\end{document}